\documentclass{article}

\usepackage{biblatex}
\usepackage{amsmath}
\usepackage{amssymb}
\usepackage{subcaption}
\usepackage{graphicx} 

\title{Balancing the AI Strength of Roles in Self-Play Training with Regret Matching$^+$\footnote{https://github.com/vshallc/rm-opponent.git}}
\author{Xiaoxi Wang \\ xiaoxiwang@tencent.com}
\date{}

\begin{document}

\maketitle

\begin{abstract}
When training artificial intelligence for games encompassing multiple roles, the development of a generalized model capable of controlling any character within the game presents a viable option. This strategy not only conserves computational resources and time during the training phase but also reduces resource requirements during deployment. training such a generalized model often encounters challenges related to uneven capabilities when controlling different roles. A simple method is introduced based on Regret Matching $^+$, which facilitates a more balanced performance of strength by the model when controlling various roles.
\end{abstract}

\section{Introduction}
The Self-Play algorithm \cite{silver2016mastering,tesauro1995temporal} has gained significant prominence in reinforcement learning due to its efficacy in enhancing the strength of intelligence agents in adversarial games. In competitive games with different roles (e.g. characters, factions, etc) with unique play styles and abilities to choose from, such as StarCraft with three "races" or the King of Fighters with dozens of characters, the development of a general model for intelligent agents that can play all roles is more data-efficient and cost saving than creating numerous specialized models. Training different roles on the same model, allows these roles to share their feature and action spaces and portions of the model weights; however, despite being controlled by the same model, the strength exhibited by each role can differ considerably. 

This discrepancy can be attributed to two main reasons: 1) the roles may not be designed perfectly balanced; 2) the procedure of strategy learning is related to the complexity of the roles' abilities. A role that is relatively stronger or has less complicated moves may take advantage of training convergence. Once a role $A$ has learned a dominated strategy to role $B$, in the following self-play iteration, the probability of $B$ can explore a counter strategy becomes low, since the data sampled from self-play for training $B$ are almost weak to $A$.

To avoid the incidents above, a regret matching-based data manipulating algorithm is presented in the following sections. 

\section{Training with Self-play in Competitive Games}

Let $I={1,2,...}$ denote the set of roles a player can choose in a game, and $|I|=N$. Given two roles $i, j \in I$, where $i$ is controlled by the model of the $t$th iteration and $j$ the $(t-1)$th, the probability of $i$ beats $j$ is $p(win_{i>j})$. The role $i$ is stronger than $j$ if $p(win_{i>j}) > p(win_{j>i})$ and $p(win_{i>j}) > 0.5$ if there is no draw in the game.

Let $t \in \mathcal{T} = \{0, 1, ..., T\}$ denote the $t$th iteration in a self-play training and $\theta_t$ denote the model parameter after $t$ iterations, where $t=0$ is the initial state and $\theta_0$ represents the untrained model. Each iteration contains a number of training steps, $S_t={1, 2, ...}$ and $s_t\in S_t$ denotes the $s$th training step in iteration $t$. Then the win rate of $i$ against $j$ at step $s_t$ is $p(win_{i>j}|\theta_{s_t})$, or $p_{s_t}(win_{i>j})$ for the sake of simplicity.

When training a generalized model, data samples are generated from all combinations of the roles. Typically, the proportion or weight, $w_{s_t}(i,j)$, of the sampled data for each combination is uniformly set to be $\frac{1}{N^2}$. The win rate of model at step $s_t$, $\theta_{s_t}$ against $\theta_{t-1}$ across all role combinations can be expressed as
\[
p_{s_t}(win_{all}) = \sum_{i,j\in I}p_{s_t}(win_{i>j}) w_{s_t}(i,j)
\]. In practical scenarios, even when the overall win rate converges, such as $p_{s_t}(win_{all}) \ge 1-\epsilon$ , there may still be a set of $n$ combinations for which $1 - p_{s_t}(win_{i>j}) \gg \epsilon$ if $N^2 \gg n$. For instance, in a game of $10$ roles, there are $10^2=100$ combinations, and the overall win rate could still be $0.9$ if $90$ combinations exhibit win rates of $1.0$, while the remaining $10$ never win during the training process. This outcome would indicate a failure in generalization.

\section{Manipulating the Data Distribution with Regret Matching$^+$}

In order to train an impeccable general model, the basic idea of our algorithm is to raise the weight of training data where the model's performance is subpar. Suppose there is a master player on the self-play side, who can only manipulate the weight $w_{s_t}(i,j)$ to achieve a higher win rate for its side. Consequently, the weaker the model $\theta_{s_t}$ performs on a specific combination, the greater the volume of the training data that will be utilized. 

Regret Matching (RM) and Regret Matching$^+$ (RM$^+$) \cite{hart2000simple, tammelin2014solving} method allow agents to randomly select an action proportional to the positive regrets. By applying RM$^+$ to the "master player", it enables the player to adapt the data distribution in response to the strategy employed by its opponent (the training side).

Let $R$ denote the regret matrix of size $N \times N$, where $R(i, j)$ represents the accumulated regret value of the match between the role $i$ on the training side and the role $j$ on the self-play side. It is important to note that the regret matrix $R(i, j)$ here only indicates the regret of the self-play side (role $j$). $R$ is initialized by filling a value of $\frac{1}{N^2}$ to each element.

Given a match between $i$ and $j$, the return to $j$ is defined as:

\[
    r_{s_t}(i, j)= \begin{cases}
        1, j\ \text{wins} \\
        0, \text{otherwise} \\
    \end{cases}
\]

Once the result of the match comes out, exponential smoothing is applied to the return value and this averaged result serves as an updated win rate:

\[
    p_{s_t}(win_{i>j}) = \bar{r}_{s_t}(i, j) = \bar{r}_{s_t-1}(i, j) * \gamma + r_{s_t}(i, j) * (1-\gamma)
\]

where $\gamma\in [0, 1]$ is the smoothing factor.

Then the expected utility is defined as the overall win rate, which can be calculated as: 
\[
\mathbb{E}(p_{s_t}) = p_{s_t}(win_{all}) = \sum_{i,j\in I}p_{s_t}(win_{i>j}) w_{s_t-1}(i,j)
\]

The regret is defined as the difference between the smoothed win rate and the expected win rate:
\[
    \Delta p_{s_t}(i, j) = \bar{r}_{s_t}(i, j) - \mathbb{E}(p_{s_t})
\]
and the regret matrix is updated as:
\[
    R_{s_t}(i, j) = \max(R_{s_t-1}(i, j) + \Delta p_{s_t}(i, j), 0)
\]
Finally, new weights for selecting combinations in next match is updated as:
\[
w_{s_t}(i,j) = \begin{cases}
    \frac{1}{N^2}, \text{if}\ \sum_{i, j \in I} R_{s_t}(i,j) =0, \\
    \frac{R_{s_t}(i, j)}{\sum_{i, j \in I} R_{s_t}(i,j)} * (1-\eta) + \frac{1}{N^2} * \eta \  \text{otherwise} \\
\end{cases}
\]
where $\eta$ is a weight factor.

\section{An Example}

For a game with $N=3$ roles, the regret matrix $R_0$, weights $w_0$ and win rates $p_0$ are initialized as follow:

\[
    R_0 = \begin{bmatrix}
    0.01 & 0.01 & 0.01 \\
    0.01 & 0.01 & 0.01 \\
    0.01 & 0.01 & 0.01 \\
    \end{bmatrix}
,
    w_0 = \begin{bmatrix}
        \frac{1}{9} & \frac{1}{9} & \frac{1}{9} \\
        \frac{1}{9} & \frac{1}{9} & \frac{1}{9} \\
        \frac{1}{9} & \frac{1}{9} & \frac{1}{9} \\
    \end{bmatrix}
\text{and}\ 
    p_0 = \begin{bmatrix}
        0.5 & 0.5 & 0.5 \\
        0.5 & 0.5 & 0.5 \\
        0.5 & 0.5 & 0.5 \\
    \end{bmatrix}
\]
When role $j=2$ beats $i=1$, $r_1(1, 2)=1$.
Let $\gamma=0.9$, 
\[
p_1(win_{i>j}) = \bar{r}_1(1, 2)=p_1(win_{1>2}) * 0.9 + r_1(1, 2) * 0.1 = 0.55
\]
and $p1$ is updated as:
\[
    p_1 = \begin{bmatrix}
        0.5 & 0.5 & 0.5 \\
        0.5 & 0.5 & 0.55 \\
        0.5 & 0.5 & 0.5 \\
    \end{bmatrix}
\]
Then the updated expected utility is:
\[
    \mathbb{E}(p_{1}) = p_{1}(win_{all}) = \sum_{i,j\in I}p_{1}(win_{i>j}) w_{1}(i,j)=0.5056
\]
and the regret is updated as:
\[
\Delta p_{1}(i, j) = \bar{r}_{1}(i, j) - \mathbb{E}(p_{1})=0.55-0.5056=0.0444
\]
\[
R_{1}= \begin{bmatrix}
    0 & 0 & 0 \\
    0 & 0 & 0.0444 \\
    0 & 0 & 0 \\
    \end{bmatrix}
\]
It is important to note that if $R_{1}$ is directly used as weights, then only the combination of $(1,2)$ can be selected in next round. This is the rationale behind the use of a weight factor.

As a result, the weight for next round is ($\eta=0.1$):
\[
w_1 = \begin{bmatrix}
        0.0111 & 0.0111 & 0.0111 \\
        0.0111 & 0.0111 & 0.0400 \\
        0.0111 & 0.0111 & 0.0111 \\
    \end{bmatrix}
\]

\section{Evaluation}

The evaluation is employed on a fighting game of 13 characters. Two models are trained through 10 iterations for self-play. One is vanilla self-play while the other is self-play with regret matching $^+$. On each iteration, the model continues training until convergence is achieved. Table \ref{tab:sp_rm} shows the win rate among all role combinations. 
\begin{table}[h]
\begin{subtable}[t]{0.45\textwidth}
\begin{tabular}{|l||l|l|l|l|l|l|l|l|l|l|l|l|l|}
\hline
Role ID & 1    & 2    & 3    & 4    & 5    & 6    & 7    & 8    & 9    & 10   & 11   & 12   & 13   \\ \hline\hline
1       & 0.3  & 0.1  & 0.08 & 0.3  & 0.33 & 0.07 & 0.1  & 0.18 & 0.02 & 0.03 & 0.16 & 0    & 0.32 \\ \hline
2       & 1    & 0.43 & 0.35 & 0.8  & 0.93 & 0.96 & 0.05 & 0.88 & 0.68 & 0.27 & 0.88 & 0.18 & 0.37 \\ \hline
3       & 0.73 & 0.53 & 0.6  & 0.36 & 0.16 & 0    & 0.38 & 0.97 & 0.5  & 0.6  & 0.93 & 0.2  & 0.2  \\ \hline
4       & 0.88 & 0.43 & 0.52 & 0.5  & 0.15 & 0.13 & 0.1  & 0.26 & 0.2  & 0.08 & 0.17 & 0.45 & 0.23 \\ \hline
5       & 0.72 & 0.05 & 0.9  & 0.87 & 0.44 & 0.49 & 0.63 & 0.9  & 0.11 & 0.91 & 0.67 & 0.56 & 0.5  \\ \hline
6       & 0.87 & 0.06 & 0.88 & 0.84 & 0.69 & 0.6  & 0.65 & 0.86 & 0.18 & 0.85 & 0.84 & 0.09 & 0.4  \\ \hline
7       & 0.8  & 0.77 & 0.67 & 0.95 & 0.5  & 0.5  & 0.43 & 0.96 & 0.34 & 0.28 & 1    & 0.3  & 0.73 \\ \hline
8       & 0.78 & 0.06 & 0    & 0.5  & 0.1  & 0.18 & 0.05 & 0.53 & 0.77 & 0.98 & 0.05 & 0.04 & 0.4  \\ \hline
9       & 0.96 & 0.37 & 0.2  & 0.76 & 0.93 & 0.72 & 0.71 & 0.2  & 0.57 & 0.75 & 0.68 & 0.02 & 0.75 \\ \hline
10      & 0.92 & 0.67 & 0.33 & 1    & 0.12 & 0.04 & 0.65 & 0.1  & 0.34 & 0.6  & 0.53 & 0.03 & 0.33 \\ \hline
11      & 0.89 & 0.2  & 0.09 & 0.8  & 0.16 & 0.2  & 0    & 1    & 0.37 & 0.55 & 0.5  & 0.12 & 0.77 \\ \hline
12      & 1    & 0.76 & 0.8  & 0.53 & 0.33 & 0.82 & 0.72 & 0.93 & 1    & 0.96 & 0.8  & 0.5  & 0.24 \\ \hline
13      & 0.8  & 0.67 & 0.77 & 0.72 & 0.6  & 0.4  & 0.28 & 0.72 & 0.4  & 0.62 & 0.3  & 0.76 & 0.38 \\ \hline
\end{tabular}
\caption{Self-play, no regret matching}
\label{tab:sp_wo_rm}
\end{subtable}
\newline
\vspace{0.5cm}
\newline
\begin{subtable}[t]{0.45\textwidth}
\begin{tabular}{|l|l|l|l|l|l|l|l|l|l|l|l|l|l|}
\hline
Role ID & 1    & 2    & 3    & 4    & 5    & 6    & 7    & 8    & 9    & 10   & 11   & 12   & 13   \\ \hline
1       & 0.45 & 0.88 & 0.62 & 0.93 & 0.74 & 0.32 & 0.59 & 0.76 & 0.68 & 0.16 & 0.8  & 0.37 & 0.18 \\ \hline
2       & 0.39 & 0.53 & 0.24 & 0.39 & 0.8  & 0.64 & 0.77 & 0.74 & 0.61 & 0.8  & 0.66 & 0.31 & 0.68 \\ \hline
3       & 0.54 & 0.66 & 0.55 & 0.59 & 0.47 & 0.67 & 0.63 & 0.46 & 0.77 & 0.5  & 0.66 & 0.76 & 0.37 \\ \hline
4       & 0.04 & 0.71 & 0.64 & 0.63 & 0.7  & 0.25 & 0.12 & 0.49 & 0.47 & 0.06 & 0.09 & 0.67 & 0.22 \\ \hline
5       & 0.3  & 0.32 & 0.51 & 0.3  & 0.53 & 0.21 & 0.15 & 0.43 & 0.24 & 0.12 & 0.29 & 0.31 & 0.26 \\ \hline
6       & 0.68 & 0.29 & 0.34 & 0.56 & 0.71 & 0.52 & 0.78 & 0.81 & 0.81 & 0.9  & 0.53 & 0.73 & 0.87 \\ \hline
7       & 0.31 & 0.23 & 0.24 & 0.78 & 0.88 & 0.39 & 0.48 & 0.5  & 0.74 & 0.46 & 0.39 & 0.42 & 0.14 \\ \hline
8       & 0.19 & 0.35 & 0.67 & 0.48 & 0.5  & 0.22 & 0.4  & 0.42 & 0.73 & 0.43 & 0.69 & 0.68 & 0.24 \\ \hline
9       & 0.37 & 0.41 & 0.3  & 0.51 & 0.62 & 0.09 & 0.42 & 0.56 & 0.54 & 0.45 & 0.63 & 0.11 & 0.4  \\ \hline
10      & 0.7  & 0.29 & 0.5  & 0.88 & 1    & 0.14 & 0.5  & 0.55 & 0.6  & 0.33 & 0.85 & 0.74 & 0.16 \\ \hline
11      & 0.31 & 0.4  & 0.32 & 0.8  & 0.56 & 0.36 & 0.61 & 0.5  & 0.2  & 0.05 & 0.53 & 0.23 & 0.3  \\ \hline
12      & 0.72 & 0.55 & 0.24 & 0.33 & 0.73 & 0.31 & 0.54 & 0.36 & 0.86 & 0.43 & 0.88 & 0.68 & 0.15 \\ \hline
13      & 0.91 & 0.37 & 0.7  & 0.69 & 0.62 & 0.17 & 0.87 & 0.59 & 0.68 & 0.81 & 0.79 & 0.88 & 0.47 \\ \hline
\end{tabular}
\caption{Self-play with regret matching}
\label{tab:sp_w_rm}
\end{subtable}
\caption{The win rates on all role combinations for model with/without regret matching $^+$}
\label{tab:sp_rm}
\end{table}
The variance of table \ref{tab:sp_wo_rm} is $0.0964$, while of table \ref{tab:sp_w_rm} is $0.0554$.

\printbibliography
\end{document}